\pdfoutput=1
\documentclass[10pt]{article}
\usepackage{dmstyle}

\usepackage{amsmath,amsfonts,bm}

\def\eqref#1{equation~\ref{#1}}

\def\1{\bm{1}}

\DeclareMathAlphabet{\mathsfit}{\encodingdefault}{\sfdefault}{m}{sl}
\SetMathAlphabet{\mathsfit}{bold}{\encodingdefault}{\sfdefault}{bx}{n}

\newcommand{\E}{\mathbb{E}}

\newcommand{\R}{\mathbb{R}}

\usepackage{hyperref}
\usepackage{url}
\usepackage{amsmath,amssymb,booktabs,graphicx,multirow}
\usepackage{placeins}
\usepackage{wrapfig,capt-of}
\usepackage{xcolor}
\newcommand{\rev}[1]{#1}

\newcommand{\model}{AllocFlow}
\newcommand{\one}{\mathbf{1}}
\newcommand{\sg}{\operatorname{sg}}
\newcommand{\clip}{\operatorname{clip}}

\title{AllocFlow: End-to-End Portfolio Trading Action Generation from Raw Market Data}
\shorttitle{AllocFlow: End-to-End Portfolio Trading Action Generation from Raw Market Data}

\author{Annan Wang$^{1,*}$, Yuchen Xie$^{1,*}$, Yu Cen$^{2}$, Leon Liangyu Chen$^{3}$,
  Qianyu Julie Zhu$^{4}$, Xin Li$^{1}$, Tianyi Wei$^{1}$ and Weisi Lin$^{1}$}

\affiliations{\mbox{$^{1}$Nanyang Technological University}, \mbox{$^{2}$Independent Researcher},
  \mbox{$^{3}$Stanford University}, \mbox{$^{4}$Massachusetts Institute of Technology}\\
  $^{*}$Equal contribution}

\projectpage{https://allocflow.github.io}
\contact{annan001@e.ntu.edu.sg}

\logos{%
  \logo[1.15cm]{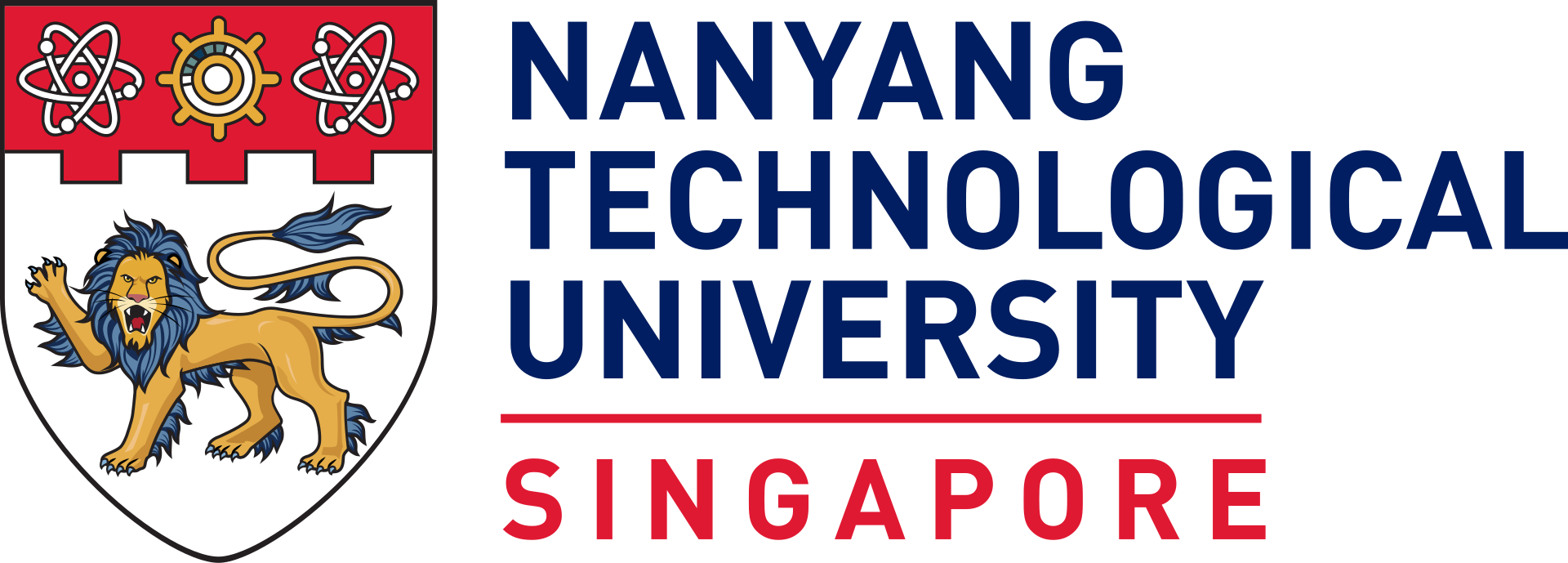}\logosep
  \logo[0.70cm]{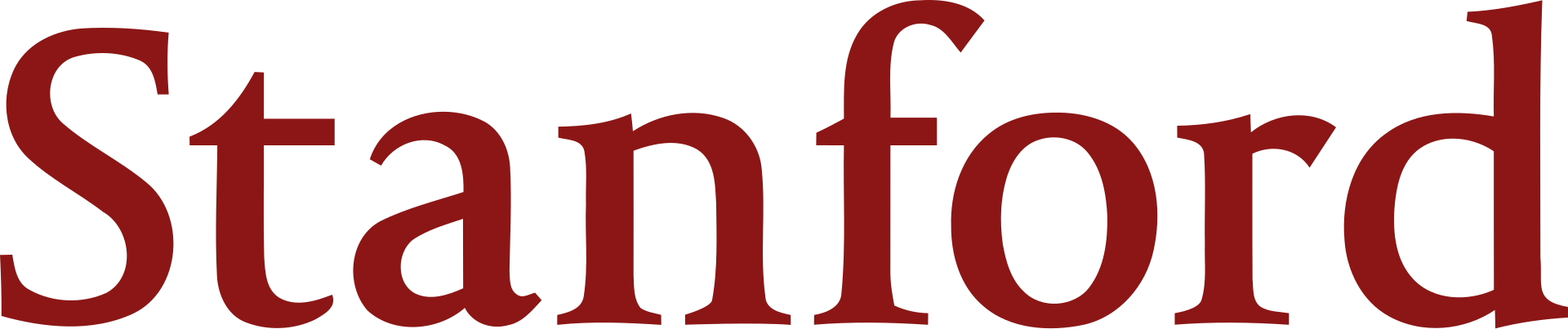}\logosep
  \logo[0.95cm]{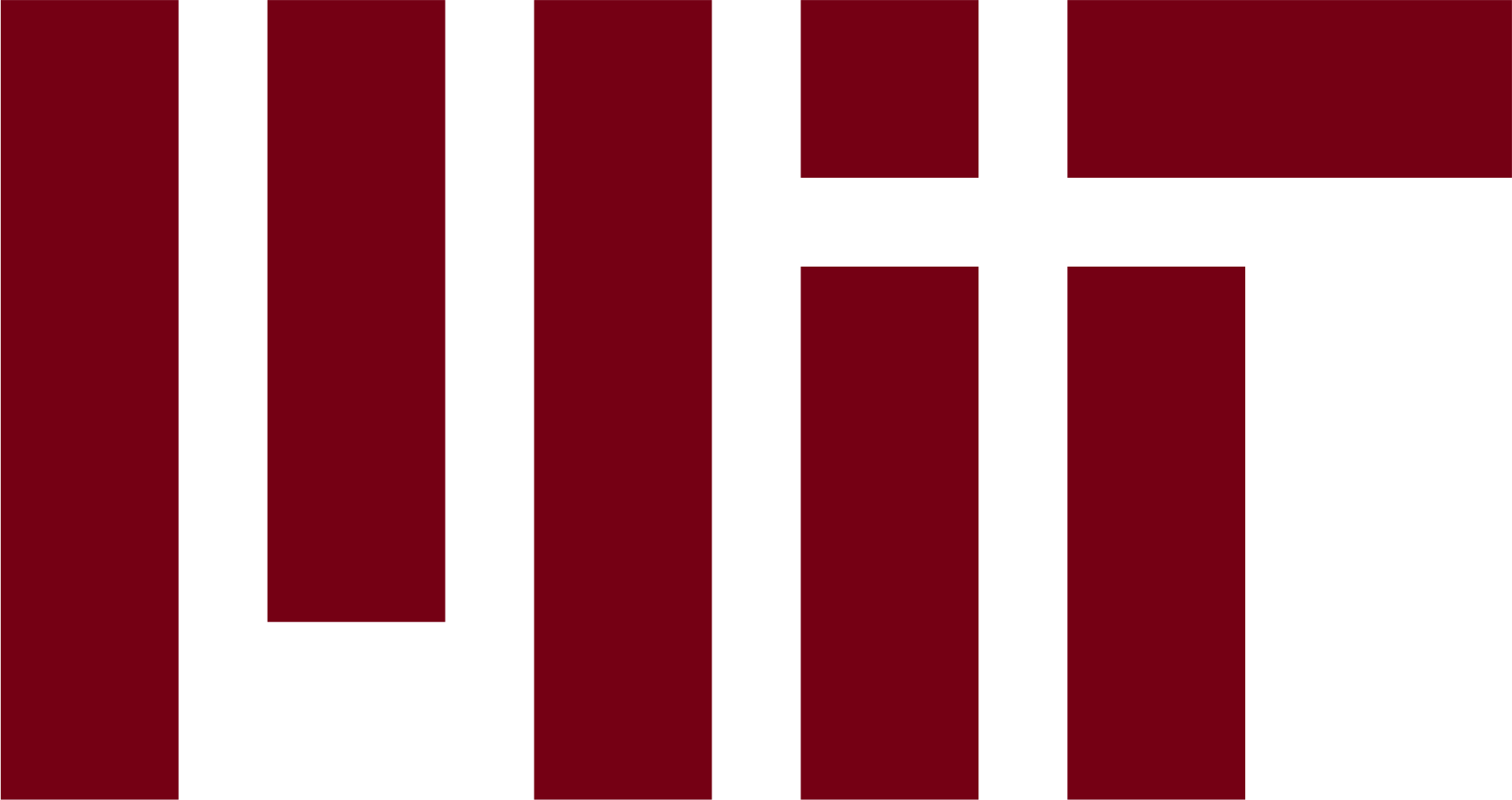}}

\begin{document}
\begin{abstract}
Real-world portfolio management requires converting noisy market observations into coordinated, simultaneous allocations across multiple assets under shared capital constraints. Advances in forecasting models and alpha discovery help identify opportunities. However, traditional pipelines often decouple signal forecasting from portfolio construction, potentially misaligning learning with allocation and execution objectives. We present AllocFlow, an end-to-end flow-matching framework that directly generates joint portfolio allocations from raw multi-asset market histories. To capture dependencies among portfolio positions, AllocFlow employs a cross-asset flow Transformer conditioned on temporal market representations to generate continuous, signed position weights across all instruments jointly. To address the sample inefficiency and instability of learning continuous allocation policies from scratch, we establish a two-stage training scheme: strategy-supervised imitation shapes a structured distribution of candidate portfolios, followed by group-relative reinforcement learning (RL) that refines joint decisions using portfolio-level financial rewards and a frozen policy reference. \rev{On hourly cryptocurrency perpetual futures, RL refinement raises the annualized net return of the imitation policy from $27.79\%$ to $51.38\%$ and its Calmar ratio from $0.955$ to $4.043$, surpassing the teacher strategy on every metric, and AllocFlow reaches a Calmar ratio of $4.90$ on six further months of unseen 2026 data without retraining.}
\end{abstract}
\maketitle

\begin{figure}[!ht]
\centering
\includegraphics[width=\linewidth]{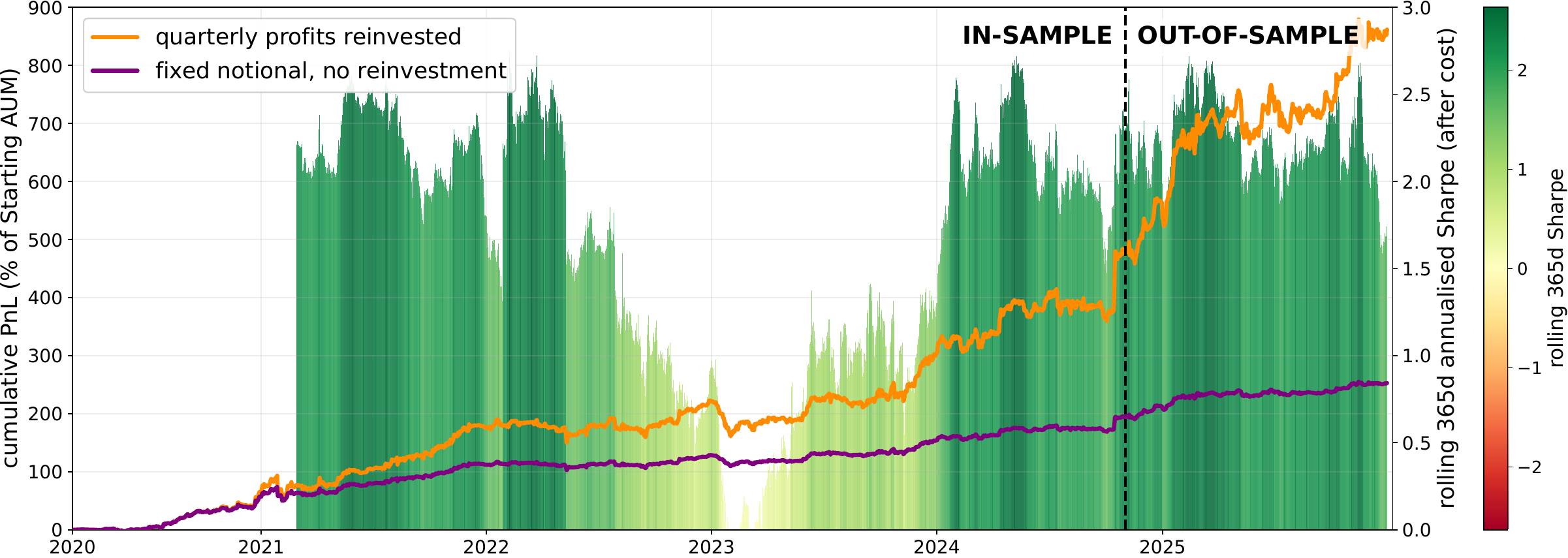}
\caption{\textbf{After-cost trading performance} from 01 Jan 2020 to 31 Dec 2025. The dashed line marks the out-of-sample start. Purple shows cumulative PnL on fixed notional, orange shows a view with profits reinvested quarterly, and color indicates rolling 365-day annualized Sharpe.}
\label{fig:teaser}
\end{figure}

\section{Introduction}
Algorithmic trading has evolved from explicit rules and statistical portfolio construction toward learned signals, policies, and market representations. Classical portfolio construction formalizes the balance between expected return and risk \citep{markowitz1952portfolio}, while modern machine learning extracts richer predictive signals \citep{gu2020empirical}. Recent financial foundation models learn reusable representations of price histories \citep{shi2025kronos}, and language-model agents and reinforcement learning (RL) increasingly automate alpha discovery \citep{yu2023alphagen,tang2025alphaagent,li2025rdagent,chen2026alphasage}. Yet an improved forecast or alpha does not by itself determine a better trade. Its economic value depends on how it is converted into positions, combined with other opportunities, and implemented through rebalancing. Optimizing these intermediate signals separately can therefore leave a gap between what a model learns and what a trading account needs to achieve \citep{elmachtoub2022spo}.

An account must choose \emph{how much to allocate to each instrument, in which direction}, under shared capital and exposure considerations. A correct directional forecast can still lead to a poor decision if the position is oversized, duplicates another exposure, or costs too much to change. Conversely, coordinated long and short positions can exploit relative opportunities without relying on a large directional market bet. Each position must therefore be considered in relation to the others. Direct trading and portfolio policies have a substantial history \citep{moody2001direct,jiang2017portfolio,li2023lsrecaan}. We investigate how to realize this joint decision through an end-to-end generative model and a training procedure that produces profitable portfolio behavior.

Our goal is to make joint trading decisions for the entire portfolio directly from raw market observations. To realize this challenging goal, the action representation must express coordinated positions, while training must discover profitable allocations within a large continuous space. Historical observations provide noisy outcomes and no unique optimal portfolio label. Temporal pretraining supplies useful market representations \citep{zhou2023onefitsall,jin2024timellm,shi2025kronos}, but does not itself determine a trading policy. RL can optimize financial outcomes directly \citep{moody2001direct,deng2016deep,zhang2020deep}; however, uninformed exploration must discover meaningful position relationships and sizing while learning from noisy returns. Trading costs can further favor inactivity when candidate portfolios have insufficient expected benefit. Financial RL frameworks have tried to address parts of these challenges \citep{liu2020finrl,jiang2024reinforcement}, and generative policies offer expressive continuous-action models \citep{chi2023diffusionpolicy}. The remaining design question is how to combine an appropriate joint action model with an informative starting point and effective portfolio-level optimization.

We propose \model, a cross-asset flow Transformer that generates continuous, signed position weights conditioned on pretrained temporal market representations. Together with the model, we designed a two-stage training recipe that separates acquiring trading structure from improving portfolio outcomes. We train the randomly initialized action head through imitation learning (IL) of an alpha-based strategy, teaching relative position preferences, direction, and scale to yield a structured distribution of candidate portfolios. This gives RL a coherent starting region for search without making the teacher's standalone performance the final objective. Group-relative RL then compares candidate portfolios under shared market outcomes and improves their financial rewards, while a frozen policy reference guides the updates. The encoder supplies market information; the action module learns how to convert that information into coordinated allocations and refine them through reward feedback. \rev{Here, \emph{end-to-end} means that \model\ maps raw market histories to executable target weights with no intermediate forecast, alpha score, or optimizer; the pretrained encoder stays frozen, as in end-to-end vision-language and robot policies \citep{liu2023llava,shridhar2021cliport,nvidia2025gr00t}, and allocations are learned directly under a portfolio objective \citep{zhang2020dlportfolio}. We focus on cryptocurrency perpetual futures, which trade around the clock, allow long and short positions at the same fee, and offer a liquid, weekly rotating universe.}

\rev{On hourly cryptocurrency perpetual futures, RL refinement raises the annualized net return from $27.79\%$ after imitation to $51.38\%$, the Sharpe ratio from $0.954$ to $2.192$, and the Calmar ratio (annualized net return over maximum drawdown) from $0.955$ to $4.043$ over 14 months, so that \model\ surpasses its own teacher on every metric; it also reaches a Calmar of $4.90$ on six further months of 2026 data without retraining. Each part of the design is necessary: under the identical RL objective, a parameter-matched Gaussian policy reaches a Calmar of at most $1.92$, direct differentiable optimization of the same reward at most $1.74$, and RL from a randomly initialized head loses $75\%$ per year.}

Our contributions are threefold: (i) \textbf{Joint action generation from market histories.} A frozen financial encoder compresses each asset's history, and a cross-asset flow Transformer produces the signed target book \rev{(the vector of signed position weights)} in one decision; (ii) \textbf{Strategy-guided refinement.} Imitation supplies coherent portfolio proposals, then group-relative RL with a frozen reference improves their portfolio-level reward; (iii) \rev{\textbf{Strong performance beyond the teacher.} RL raises the Calmar of the imitation policy from $0.955$ to $4.043$, more than twice that of the teacher ($1.733$), and replacing the action head, encoder, teacher, or reward, or removing imitation, each yields a lower Calmar.}

\section{Related Works}
\paragraph{From market information to decisions.}
Forecasting and financial foundation models learn market representations \citep{gu2020empirical,shi2025kronos,liu2023itransformer,song2026label}, while AlphaGen, AlphaSAGE, AlphaAgent, and R\&D-Agent-Quant mine predictive signals \citep{yu2023alphagen,chen2026alphasage,tang2025alphaagent,li2025rdagent}. Their outputs can inform a portfolio, but require a further rule to choose signed positions. \model\ instead learns the complete joint allocation and evaluates its realized outcome.

\paragraph{Direct portfolio learning.}
Trading and portfolio RL already map observations to actions \citep{moody2001direct,jiang2017portfolio,li2023lsrecaan,zhang2020dlportfolio}, and imitation-assisted portfolio learning is established \citep{dong2024sirl}. \rev{Recent RL portfolio methods adapt to risk and market regimes \citep{chen2026mars,kim2025deeparies,pan2026recap}, and crypto trading agents are guided by hindsight-optimal oracles \citep{qin2024earnhft,zong2024macrohft,qin2026fineft}.} Recent approaches combine learned information with mean-variance allocation, executable programs, Black-Litterman scenarios, differentiable sparse optimization, or ranked allocation \citep{yu2024fincon,song2026timi,soun2026stable,jeon2026sparse,zhang2025ram}. \rev{Our contribution is a flow-generated continuous joint action and a strategy-guided refinement procedure.}
The operative difference is what the policy emits: an alpha score, an optimizer input, a trading program, and a vector of executable target weights define different decision boundaries. Our experiments therefore compare complete deployed portfolios while using within-model controls to examine the action generator itself.

\paragraph{Generative action refinement.}
Flow matching transports noise to continuous samples \citep{lipman2022flow}; diffusion and flow policies apply generative models to actions \citep{chi2023diffusionpolicy,black2024pi0}. DPPO, Flow-GRPO, and ReinFlow demonstrate reinforcement updates for such policies \citep{ren2025dppo,liu2025flowgrpo,zhang2025reinflow}. These methods are widely adopted in continuous-space dense generation tasks, e.g., image generation and robotic action generation. We adapt this connection to portfolio trading actions, using imitation for initialization and a frozen reference during reward refinement. Appendix~\ref{app:related} discusses execution-aware trading and other generative finance work.

\section{Method}
\label{sec:method}
\begin{figure}[ht]
\centering
\includegraphics[width=\linewidth]{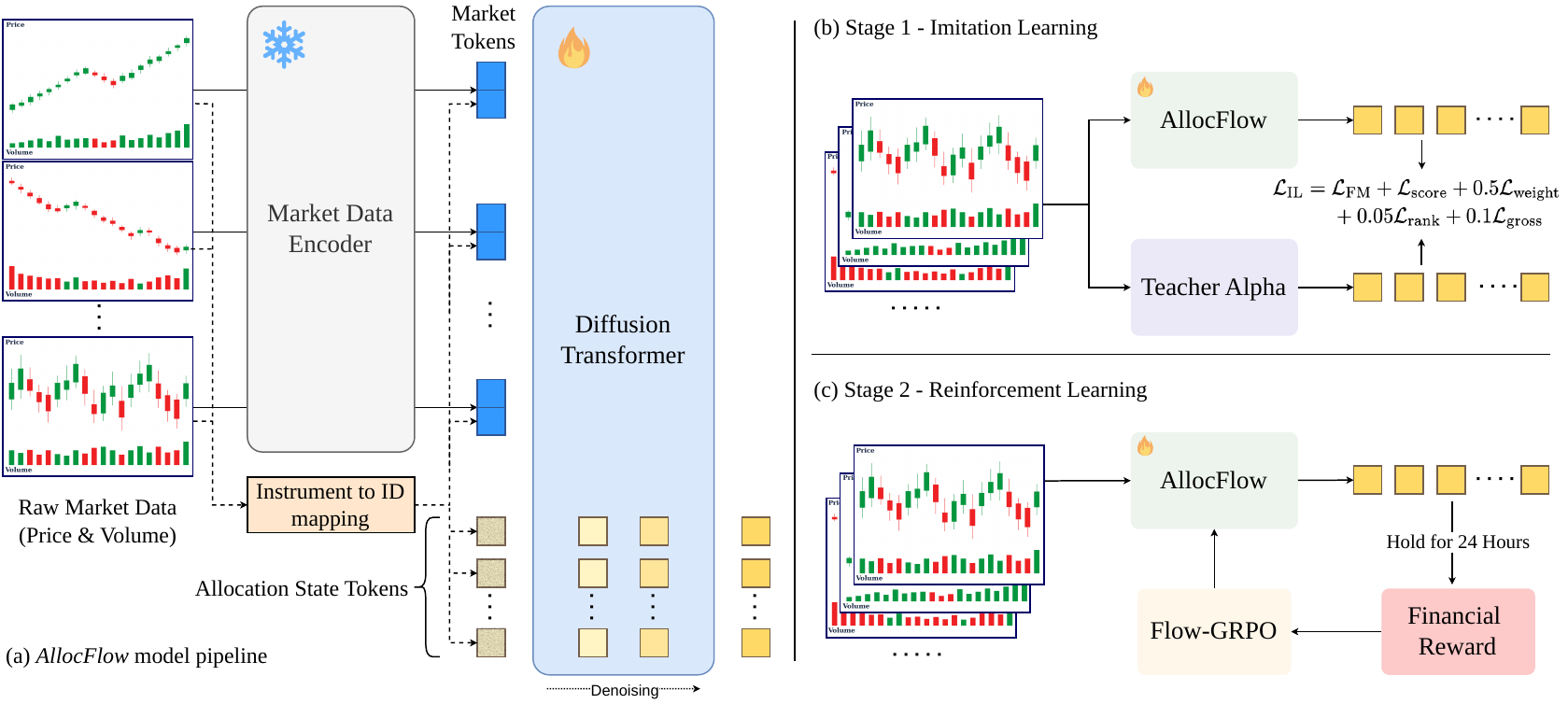}
\caption{\textbf{\model\ from market histories to portfolio decisions.} (a) Frozen Kronos yields a coarse/fine market-token pair per asset; the instrument-ID mapping connects each pair to its allocation-state token. The DiT-style flow head generates a joint target book. (b) Causal strategy allocations initialize the head through imitation. (c) A 24-hour portfolio reward and a frozen IL reference guide Flow-GRPO refinement.}
\label{fig:overview}
\end{figure}

A market representation can provide a compact market context without deciding how it should be traded alongside other assets. \model\ aims to generate a complete target portfolio from raw market histories. As Figure~\ref{fig:overview} shows, a frozen encoder first condenses those histories, imitation gives the joint action generator a useful starting distribution, and reward-guided refinement improves the resulting portfolio decisions.

\subsection{Generating a portfolio from compact market context}
Portfolio weights should be decided together: a position's value depends on the other exposures and the neutrality of the complete book. \rev{Market data arrive as hourly \emph{bars}, each aggregating the open, high, low, and close prices and the traded volume (OHLCV) over one hour; Appendix~\ref{app:terms} defines all financial terms.} At market time $t$, let $\mathcal U_t$ contain $N_t$ eligible assets and let $X_t\in\R^{N_t\times L\times C}$ contain their available histories. The policy produces signed target weights $w_t\in\R^{N_t}$ relative to a fixed account notional. Positive and negative weights denote long and short positions. Execution rounds the target to $\widehat w_t$ and applies it after one bar. In the fixed-notional ledger, the trading action is simply $\Delta w_{t+1}=\widehat w_t-\widehat w_{t-1}$, with zero weight assigned to names outside each target's asset list. Thus the model learns the desired book, while the ledger obtains orders and fees from changes to that book (Appendix~\ref{app:metrics}).

Direct attention over every bar of every asset would make the action head's context long. We therefore freeze Kronos \citep{shi2025kronos}\rev{, a financial foundation model pretrained to predict the next bar from tokenized OHLCV sequences,} and use its temporal representation rather than its next-bar forecast. For each asset, normalized OHLCV and derived amount become coarse and fine token IDs. A single forward pass provides two hidden states before their logit heads: the final normalized coarse-decoder context and the fine dependency-layer output conditioned on that context and a coarse-token query. We mean-pool the last eight steps of each stream, obtaining one coarse and one fine market token per asset. This path neither samples a next-bar token nor feeds a predicted price to the policy (Appendix~\ref{app:context}). \rev{Kronos is designed for forecasting rather than trading: its own forecasts lose money after fees, zero-shot and after LoRA adaptation (Table~\ref{tab:direct}), so \model\ takes only its representation and learns every trading decision in the flow head.}

We interleave the $N_t$ token pairs and append their mean as a shared market token, yielding $2N_t+1$ tokens for the DiT-style flow head \citep{peebles2023dit}. The instrument-to-ID mapping associates each pair with the allocation-state token whose output is that asset's target weight (Figure~\ref{fig:overview}a). The head can therefore compare opportunities across assets while generating their positions together. At inference, both token sets are assembled for the current $\mathcal U_t$: membership, size, and order may change for instruments with mapped IDs, without fixed asset-specific output slots. ``End-to-end'' refers to this bars-to-weights inference path; Kronos remains frozen. The frozen-encoder comparison in Table~\ref{tab:encoder} examines the representation choice.

The allocation-state tokens form a continuous action vector, so a flow model can generate alternative complete books instead of assigning positions independently. Starting from $z_0\sim\mathcal N(0,I)$, the market-conditioned velocity field evolves the vector according to $\mathrm dz_\tau/\mathrm d\tau=v_\theta(z_\tau,\tau;X_t)$. Flow time $\tau\in[0,1]$ indexes generation and is distinct from market time $t$. Each update changes the whole vector; four Euler steps are used at inference. The final state is decoded as
\begin{equation}
 w_t=\frac{1}{c}P_tz_1,\qquad P_t=I-\frac{\one\one^\top}{N_t},\qquad c=10.
 \label{eq:decode}
\end{equation}
Projection removes the common component, imposing $\one^\top w_t=0$ before execution rounding, while the model still chooses relative directions, sizes, and variable gross exposure $\|w_t\|_1$. There is no additional hard gross cap in the evaluated decoder; dollar neutrality alone does not imply beta neutrality or bounded risk (Appendix~\ref{app:neutrality}). Current holdings enter the trading ledger, not the policy input\rev{; Section~\ref{sec:objective} shows that adding them to the input extends \model\ directly to closed-loop control}. The action-head and depth ablations in Tables~\ref{tab:action} and~\ref{tab:depth} test this joint generator.

\rev{\paragraph{Why flow matching.} Flow matching suits portfolio allocation for three reasons. First, the action is a joint distribution over complete books of signed weights, for example, when several correlated assets are candidates for the same short. Second, group-relative RL needs coherent alternatives: the flow samples complete books from the learned distribution, whereas a diagonal Gaussian perturbs one mean book coordinate by coordinate. Third, the stochastic flow sampler provides the closed-form transition likelihoods that Flow-GRPO optimizes. A parameter-matched Gaussian policy trained with the identical objective reaches less than half of the flow policy's Calmar (Section~\ref{sec:heads}).}

\subsection{Imitation learning}
A continuous generator trained only from financial rewards would have to discover coordinated books through noisy outcomes and trading costs. We therefore initialize the action head randomly and train it by imitation learning on causal strategy decisions $w_t^\star$; only the frozen Kronos encoder starts from pretrained weights. We design our teacher method, named AgenticAlpha, by aggregating effective alphas mined by AlphaGen~\citep{yu2023alphagen} and AlphaSAGE~\citep{chen2026alphasage} to supply the reference demonstrations \rev{(the two alpha sets each manage half of the capital; Appendix~\ref{app:context} gives the construction and its leading alphas)}, while also employing other common teachers to test whether refinement depends on this particular teacher as discussed in Section~\ref{app:init} and Table~\ref{tab:teachers}. These are historical strategy actions, not hindsight-optimal portfolios: imitation supplies plausible directions and scale for exploration, without making the teacher's return the final objective.

Conditional flow matching moves noise toward each demonstrated book. With $a_t=cw_t^\star$, $\epsilon\sim\mathcal N(0,I)$, and sampled flow time $\tau$, define $z_\tau=(1-\tau)\epsilon+\tau a_t$ and minimize
\begin{equation}
 \mathcal L_{\mathrm{FM}}=\E\!\left[\frac{1}{N_t}\left\|v_\theta(z_\tau,\tau;X_t)-(a_t-\epsilon)\right\|_2^2\right].
 \label{eq:fm}
\end{equation}
This teaches a distribution over complete allocations. Because useful search proposals also require reliable relative preferences and nontrivial sizing, we supplement flow matching with a direct-score readout and losses on the estimated terminal action $\widehat a=z_\tau+(1-\tau)v_\theta(z_\tau,\tau;X_t)$. The imitation objective is
\begin{equation}
 \mathcal L_{\mathrm{IL}}=\mathcal L_{\mathrm{FM}}+\mathcal L_{\mathrm{score}}
 +0.5\mathcal L_{\mathrm{weight}}+0.05\mathcal L_{\mathrm{rank}}+0.1\mathcal L_{\mathrm{gross}}.
 \label{eq:il}
\end{equation}
The ranking term preserves which assets the teacher favors; the weight and gross terms preserve direction, relative size, and book scale. Appendix~\ref{app:coldstart} discusses the uninformative no-trade case. Imitation shapes where RL begins without requiring it to reproduce a teacher action. The teacher ablation in Table~\ref{tab:teachers} examines this initialization choice.

\subsection{Refining joint decisions with portfolio rewards}
Imitation makes the generated books meaningful, but it cannot tell the policy which deviations improve portfolio outcomes. Flow-GRPO~\citep{liu2025flowgrpo} therefore compares complete candidate books under the same future market path. For each historical context, the old policy samples $G=8$ candidates through $K=10$ stochastic flow transitions. Each is held fixed over the same $H=24$ forward hourly bars. To compare composition rather than reward a mere increase in leverage, we normalize each decoded candidate to gross exposure $g_0=0.85$ before scoring:
\begin{equation}
 \widetilde w_i=g_0\frac{w_i}{\|w_i\|_1+\varepsilon},\quad
 p_{i,h}=\widetilde w_i^\top r_{t,h},\quad
 R_i=\sqrt{8760}\frac{\operatorname{mean}_{h=1}^{H}p_{i,h}}
 {\operatorname{std}_{h=1}^{H}p_{i,h}+\varepsilon}.
 \label{eq:reward}
\end{equation}
Here $r_{t,h}$ is the vector of returns $h$ bars ahead. \rev{Scoring all candidates of a state on the same path at the same gross exposure makes the reward a controlled comparison of book composition. Stage~2 is thus a contextual bandit over target books: as in standard portfolio RL for liquid markets \citep{jiang2017portfolio}, the policy is a price taker whose trades do not move prices, so holdings affect outcomes only through the trading costs that the ledger charges on every change of the target book, and the reward measures the quality of a complete book without the credit-assignment noise of long, path-dependent rollouts. All reported results use the deployed after-fee ledger, where the policy sets its own exposure every hour (Section~\ref{sec:objective} compares deployment-aligned rewards).}

Financial rewards are noisy, so unrestricted updates could discard the structure learned from demonstrations. We keep a copy of the imitation-learned policy $\theta_{\mathrm{IL}}$ frozen as a soft reference while updating only the action head. Following clipped policy optimization \citep{schulman2017ppo,shao2024deepseekmath} and stochastic flow-policy refinement \citep{liu2025flowgrpo,zhang2025reinflow}, let $A_i=\clip((R_i-\overline R)/(s_R+\varepsilon),-5,5)$ be the group-relative advantage. For mean per-asset transition log density $\ell_\theta^{i,k}$, set $q_{i,k}=\exp(\ell_\theta^{i,k}-\ell_{\mathrm{old}}^{i,k})$. The update minimizes
\begin{equation}
 \mathcal L_{\mathrm{RL}}=\E\left[\mathcal C(q_{i,k},A_i)
 +\lambda\mathcal C(q_{i,k},B_{i,k})+\beta\mathcal K_{i,k}\right].
 \label{eq:rl}
\end{equation}
Here $\mathcal C$ is the negative clipped surrogate. The detached $B_{i,k}$ favors transitions scored more highly by this frozen reference than by the old policy, following the on-policy distillation idea of \citet{wang2026openclaw}; $\mathcal K_{i,k}$ penalizes deviation of transition means from that reference. We use $\lambda=0.1$, $\beta=0.004$, and ratio clipping $\delta=10^{-3}$; Appendix~\ref{app:rl} specifies the sampler and dimension normalization. The reference guides exploration without forcing exact imitation. After training, the same head generates target books with four deterministic Euler steps. The reward ablation in Table~\ref{tab:reward} examines the choice of financial objective.

\section{Experiments}
\label{sec:experiments}
The main experiment asks whether generating a joint target book improves realized, after-fee portfolio outcomes. We first compare imitation with its RL-refined continuation under the same model and ledger, then place the complete policy alongside published alpha-mining methods. Section~\ref{sec:ablations} examines which design choices account for the observed differences.

\subsection{Model Training}
\label{sec:training}
We train the randomly initialized action head in two stages while keeping the pretrained Kronos encoder frozen. First, AgenticAlpha produces a target portfolio for each historical market context using only information available at that time. The head learns from these examples through flow matching; auxiliary losses help it preserve the teacher's position sizes, asset preferences, and total exposure. This imitation learning (IL) stage gives the policy plausible complete portfolios before it is asked to improve financial outcomes.

In the second stage, the policy generates eight alternative portfolios for each training context. We scale each to $0.85$ gross exposure, hold it fixed over the same next 24 hourly bars, and compute its gross Sharpe. This controls for leverage and market path when comparing candidate allocations. Flow-GRPO compares the eight scores and updates the head toward better candidates, while a frozen copy of the imitation-learned policy keeps updates anchored. Demonstrations and reward windows exclude the test period. This fixed-book reward guides training; the backtest below evaluates changing target books after fees. At test time, four Euler steps generate each target portfolio. Appendix~\ref{app:rl} gives the sampler details.

\subsection{Evaluation design}
\label{sec:protocol}
We use \rev{cryptocurrency perpetual futures (futures contracts without expiry that can be held long or short)} because granular historical trading records are freely and publicly available over long periods. Policy training and test samples are strictly separated in time. Kronos pretraining ends in June 2024 \citep{shi2025kronos}, before the hourly test from November 2, 2024, to December 31, 2025; thus the frozen encoder has no test-period information in its weights. Every week, the eligible list is recomputed as the 20 assets with the highest cumulative trading volume in the preceding 60 days. This causal reconstitution tests a changing opportunity set rather than a list selected with knowledge of the full test period.

All targets enter a fixed-notional ledger after one bar. Orders are changes from current to target weights, including entries and exits when the eligible list changes, and incur 5 basis points \rev{(bps, $0.01\%$)} per unit of absolute traded weight. \rev{This fee is the maximum of the base-tier taker fees of major perpetual-futures venues and the fees in prior crypto trading papers, and the median half-spread of the traded contracts is only 0.24 bps (Appendix~\ref{app:costs}).} We report arithmetic annualized net return, additive maximum drawdown (MDD), net daily Sharpe, Calmar, and average daily turnover (TO), as defined in Appendix~\ref{app:metrics}. These measures jointly test profitability, downside exposure, and the trading needed to obtain it.

Published alpha-mining baselines use a common multi-asset signal-to-allocation pipeline, whereas \model\ generates weights directly. The IL-to-RL comparison keeps the frozen encoder and action-head architecture the same, making it the more direct test of reward refinement. Teacher strategies and native long-only Kronos deployment are examined separately in Section~\ref{sec:ablations}.

\subsection{Quantitative Results}
Reward refinement improves the book on both sides of the return-risk trade-off. In Table~\ref{tab:main}, net annualized return rises from $27.79\%$ after IL to $51.38\%$ after RL, while MDD falls from $29.10\%$ to $12.81\%$. The resulting Calmar increase is thus supported by both a larger numerator and a smaller drawdown, rather than by return alone. Because both stages use the same frozen representation and action-head architecture, this comparison is the main evidence that the reward stage changes the quality of the generated portfolio. The \model\ (RL) results in Table~\ref{tab:main} are means with standard deviations over \model\ trained with five random seeds; each metric is computed for each seed before averaging.

\begin{table}[ht]
\centering
\caption{\textbf{\rev{Main results.}} Return and MDD are percentages; TO is average daily turnover. Published alpha-mining baselines share a multi-asset signal-to-allocation pipeline, while \model\ generates positions directly.}
\label{tab:main}
\small
\setlength{\tabcolsep}{3.5pt} 
\begin{tabular}{lccccc}
\toprule
Policy & Return $\uparrow$ & MDD $\downarrow$ & Sharpe $\uparrow$ & Calmar $\uparrow$ & TO \\
\midrule
AlphaGen \citep{yu2023alphagen} &35.42&20.60&1.242&1.720&0.248\\
AlphaSAGE \citep{chen2026alphasage} &31.30&23.36&1.396&1.340&1.255\\
R\&D-Agent-Quant \citep{li2025rdagent} &20.69&18.53&0.967&1.117&0.397\\
AlphaAgent \citep{tang2025alphaagent} &29.12&26.93&1.048&1.081&0.651\\
\midrule
\model\ (IL) &27.79&29.10&0.954&0.955&1.143\\
\model\ (RL) & \textbf{51.38} {\scriptsize$\pm$2.04}
 & \textbf{12.81} {\scriptsize$\pm$1.51}
 & \textbf{2.192} {\scriptsize$\pm$0.086}
 & \textbf{4.043} {\scriptsize$\pm$0.379}
 & 1.976 {\scriptsize$\pm$0.047}\\
\bottomrule
\end{tabular}
\end{table}

The improvement also survives a more demanding trading profile: average daily turnover increases from $1.143$ to $1.976$, and the reported return already deducts the associated fees. Net daily Sharpe also rises, so the policy improves under a volatility-adjusted measure as well as the drawdown-based Calmar ratio. Figure~\ref{fig:teaser} adds a temporal view of one run: over a test window longer than a year, after-cost PnL trends upward and the plotted rolling 365-day Sharpe stays near 2 for much of the out-of-sample period, supporting sustained profitability. Figure~\ref{fig:reversal_cases} zooms into two concrete trading decisions changed by RL, and Section~\ref{sec:ablations} examines behavior and objective choice more directly.

Against external methods, the final policy achieves the highest reported net return and Calmar in Table~\ref{tab:main}; the strongest listed alpha baseline reaches $1.720$ Calmar. This comparison evaluates complete signal-to-position pipelines, which is relevant because \model\ is designed to decide positions jointly instead of converting independent alpha scores afterward. \rev{The fixed-backbone head comparisons in Section~\ref{sec:ablations} isolate the contribution of the action head.}

\rev{\paragraph{Longer out-of-sample test.}\label{sec:oos} Without retraining, the same policy reaches a Sharpe of $2.06$ and a Calmar of $4.90$ on six further months from January to June 2026 (Table~\ref{tab:oos} in Appendix~\ref{app:oos}), extending the out-of-sample period to 20 months, more than twice the test windows of recent RL methods for crypto trading \citep{li2025alphagat,zong2026archetypetrader}.}

\section{Ablations}
\label{sec:ablations}
The ablations study the \model\ design from three sides: market features must carry useful context, the action head must turn them into a coherent book, and reward refinement must improve on imitation. We compare heads, teachers, encoders, depth, and rewards on the after-fee ledger of Table~\ref{tab:main}; the native Kronos forecast deployment is identified separately.

\subsection{The profitability comes from the Joint Action Generation or Kronos}
\label{sec:heads}
If frozen Kronos features alone explained the portfolio gain, simpler heads on those features should benefit similarly from reward refinement. We compare the flow head with an MLP, a one-pass Transformer, and a sequential autoregressive categorical head under the reported IL and RL configurations (Table~\ref{tab:action}). The first three reach similar Calmar after imitation, but the flow head improves from $0.955$ to $4.043$ after RL, while the MLP and one-pass Transformer \rev{reach only $1.307$ and $0.960$}; the distinction emerges during refinement, which supports a distribution of complete continuous books as a match for portfolio-level search.

The categorical head supplies a further contrast: its imitation-learned book loses money, and its tested RL continuation produces no trades, \rev{consistent with poor proposals making group-relative improvement difficult. All heads share the same GRPO code and differ only in the policy kernel; the one-pass Transformer is a parameter-matched stochastic policy sampling books from $\mathcal N(\mu,\sigma^2I)$ (Appendix~\ref{app:gaussian}). Across exploration scales $\sigma$ (Table~\ref{tab:action}b), it reaches a Calmar of at most $1.92$, versus $4.043$ for the flow policy, and stays below $1$ in the other four settings. From imitation policies of similar quality (Calmar $0.878$ and $0.955$), RL thus adds $3.1$ to the flow policy's Calmar, three times its gain for the Gaussian policy.}

\begin{table}[ht]
\centering
\caption{\textbf{Action-head comparison with frozen Kronos.} (a) ``NT'' denotes an observed no-trade outcome, for which risk-adjusted ratios are not meaningful. \rev{(b) Exploration-scale sweep of the Gaussian one-pass Transformer under the identical GRPO objective; the $\sigma=0.2$ run is the one in (a).}}
\label{tab:action}
\small
\setlength{\tabcolsep}{4pt}
\begin{tabular*}{\linewidth}{@{\extracolsep{\fill}}lrrr@{\hspace{8pt}}rrr@{}}
\toprule
& \multicolumn{3}{c}{After IL} & \multicolumn{3}{c}{After RL}\\
\cmidrule(lr){2-4}\cmidrule(lr){5-7}
Head & Return (\%) & Sharpe & Calmar & Return (\%) & Sharpe & Calmar\\
\midrule
Flow Transformer &27.79&0.954&0.955&51.38&2.192&4.043\\
MLP &27.31&0.935&0.912&20.12&0.856&1.307\\
One-pass Transformer &27.78&0.949&0.878&20.85&0.786&0.960\\
Autoregressive & $-35.51$ & $-1.189$ & $-0.480$ & \multicolumn{3}{c}{NT}\\
\bottomrule
\end{tabular*}

\vspace{4pt}
{
\begin{tabular*}{\linewidth}{@{\extracolsep{\fill}}lcccccc@{}}
\multicolumn{7}{@{}l}{(b) Calmar after RL, Gaussian one-pass Transformer versus flow head}\\
\toprule
& $\sigma=0.05$ & $\sigma=0.1$ & $\sigma=0.2$ & $\sigma=0.4$ & $\sigma=0.8$ & Flow head\\
\midrule
Calmar & $-0.55$ & $-0.62$ & $0.96$ & $1.92$ & $-0.12$ & $\mathbf{4.043}$ {\scriptsize$\pm$0.379}\\
\bottomrule
\end{tabular*}}
\end{table}

Could Kronos's next-bar forecasts be traded directly, without learning a joint allocation policy? Deployed through their native long-only top-$K$ ranking rule, both zero-shot and LoRA-adapted forecasts lose money after fees (Table~\ref{tab:direct}), illustrating the gap between extracting market information and converting it into a profitable book.

\subsection{Demonstrations initialize action space; RL creates value}
\label{app:init}
Teacher variation probes the role of imitation: does it merely copy a profitable expert, or does it provide structured candidate books that RL can improve? For each of four causal strategies, AgenticAlpha (AlphaGen and AlphaSAGE signals developed on training data), MACD (12h, 26h), MACD (48h, 104h), and Alpha101 \#9 \citep{kakushadze2016alphas}, we train a new randomly initialized action head by imitation and continue it with RL on the same ledger (Table~\ref{tab:teachers}). Calmar improves in every pair, including short-horizon MACD, whose IL policy is loss-making, so the RL improvement does not require a profitable teacher.

Initialization still matters: AgenticAlpha yields the strongest refined policy among these teachers\rev{, and \model\ surpasses this teacher on every metric (Calmar $4.043$ versus $1.733$; Appendix~\ref{app:context})}. The combined pattern fits our two-stage design that demonstrations supply nontrivial directions and sizes, while reward feedback selects better portfolios from that starting region.

\rev{\paragraph{Imitation is necessary.} Flow-GRPO from a randomly initialized head, with the same encoder, sampler, reward, and RL steps (the reference terms switched off, as there is no imitation policy), produces no meaningful trading: it stays at the level of an untrained head and loses $75\%$ per year, and with $4.8$ times more RL steps, roughly the total compute of IL and RL, its gross exposure drifts to $4.5$ times the capital and its drawdown exceeds the capital. Random candidate books make reward differences mostly noise, so the group-relative advantage carries little information; imitation supplies coherent books in which this comparison becomes informative. None of 11 pure-RL trading methods trained from scratch on our data reaches the Sharpe of BTC buy-and-hold, and \model\ achieves four times the Sharpe of the best of them with one quarter of its drawdown (Appendix~\ref{app:purerl}); leading crypto RL methods likewise guide their agents with hindsight-optimal oracles \citep{qin2024earnhft,zong2024macrohft,qin2026fineft}.}

\begin{table}[ht]
\centering
\begin{minipage}[t]{0.48\linewidth}
\vspace{0pt}\centering
\captionof{table}{Kronos forecasts with native long-only deployment; these portfolios use a different allocation rule.}
\label{tab:direct}
\footnotesize\setlength{\tabcolsep}{2pt}
\begin{tabular}{lrrr}
\toprule
Kronos-direct & Return (\%) & Sharpe & Calmar\\
\midrule
Zero-shot &$-44.502$&$-0.469$&$-0.438$\\
LoRA &$-34.570$&$-0.391$&$-0.312$\\
\bottomrule
\end{tabular}
\end{minipage}\hfill
\begin{minipage}[t]{0.48\linewidth}
\vspace{0pt}\centering
\captionof{table}{Policy Calmar by teacher signal.}
\label{tab:teachers}
\footnotesize\setlength{\tabcolsep}{2pt}
\begin{tabular}{lrr}
\toprule
Teacher & After IL & After RL\\
\midrule
AgenticAlpha &0.955&4.043\\
MACD (12h, 26h) &$-0.559$&1.856\\
MACD (48h, 104h) &0.538&2.114\\
Alpha101 \#9 &0.401&1.068\\
\bottomrule
\end{tabular}
\end{minipage}
\end{table}

\begin{figure}[ht]
\centering
\includegraphics[width=\linewidth]{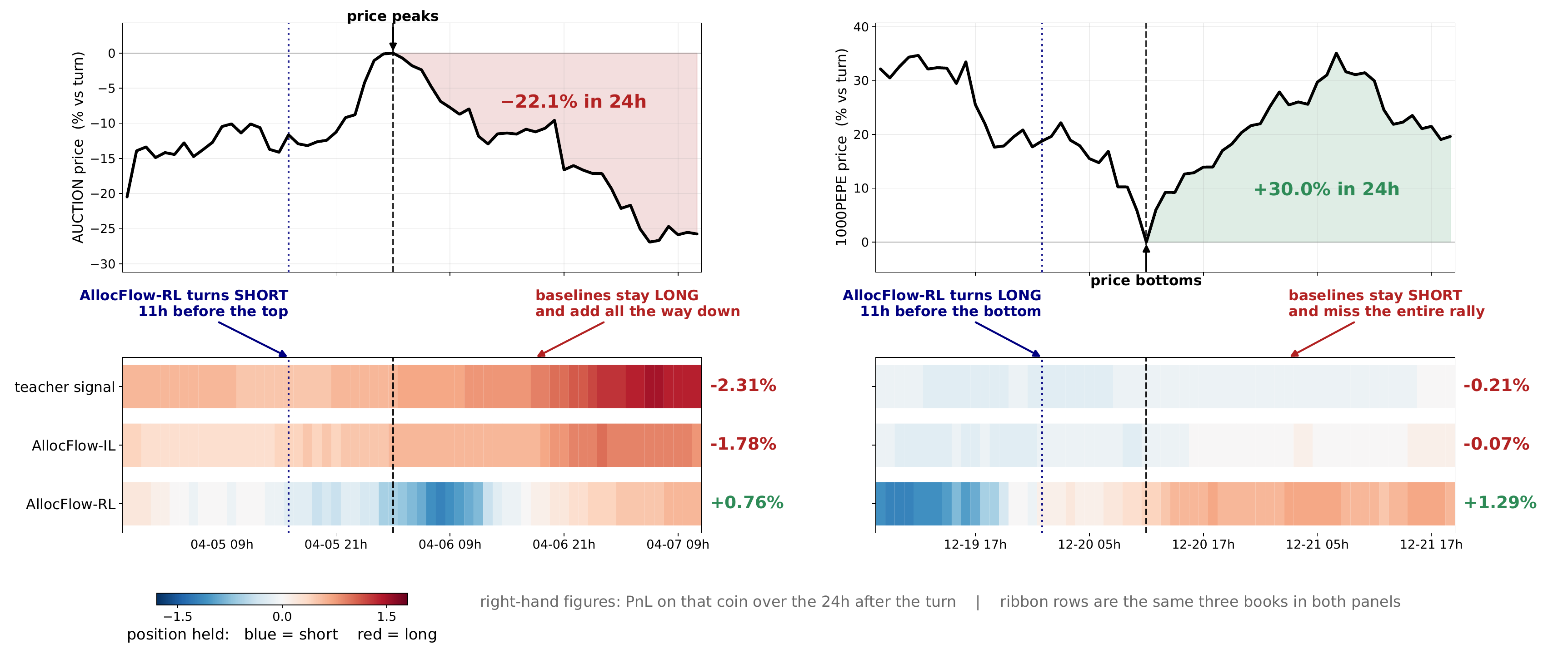}
\caption{\textbf{Opposite asset-level reversals.} Left: AUCTION falls from a local peak. Right: 1000PEPE rebounds from a local bottom. Black dashed lines mark price turns; blue dotted lines mark RL's earlier change in position direction. Blue/red positions are short/long. The labels beside each panel give asset-specific PnL over the subsequent 24 hours.}
\label{fig:reversal_cases}
\end{figure}
\FloatBarrier
\paragraph{Zoom-in to local decision traces.}
Aggregate ratios cannot reveal what changed in an individual trade, so Figure~\ref{fig:reversal_cases} aligns asset prices with the teacher, IL, and RL position ribbons around two opposite turning points: does RL change direction in time for the subsequent move, rather than merely rescale the demonstrated book? On AUCTION, RL turns short about 11 hours before the peak while the teacher and IL stay long through much of the decline; as the asset falls 22.1\% over the next 24 hours, RL records $+0.76\%$ asset-specific PnL versus $-1.78\%$ for IL. On 1000PEPE, RL covers its short and turns long about 11 hours before the bottom while the teacher and IL remain short; as the asset rises 30.0\% over the next 24 hours, RL records $+1.29\%$ versus $-0.07\%$ for IL. Refinement can thus alter the sign and timing of decisions in either direction; these windows explain a mechanism, not its frequency across the portfolio.

\subsection{Market representation}
The flow head can learn a useful book only if its market tokens preserve trading-relevant information. Replacing frozen Kronos with a frozen iTransformer while keeping the flow head loses heavily after both IL and RL on the same ledger (Table~\ref{tab:encoder}), and reward refinement does not recover the Kronos-conditioned performance. With the losing Kronos-direct deployment, this shows that good market features become valuable only when a suitable head learns to allocate from them.

\begin{table}[ht]
\centering
\begin{minipage}[t]{0.48\linewidth}
\vspace{0pt}\centering
\captionof{table}{Encoder replacement with a flow head. Return is annualized arithmetic net percent.}
\label{tab:encoder}
\footnotesize\setlength{\tabcolsep}{2pt}
\begin{tabular}{@{}lrrr@{}}
\toprule
Encoder / stage & Return & Sharpe & Calmar\\
\midrule
Kronos / IL &27.79&0.954&0.955\\
Kronos / RL &51.38&2.192&4.043\\
iTransformer / IL &$-141.58$&$-5.359$&$-0.850$\\
iTransformer / RL &$-119.29$&$-7.119$&$-0.846$\\
\bottomrule
\end{tabular}
\end{minipage}\hfill
\begin{minipage}[t]{0.48\linewidth}
\vspace{0pt}\centering
\captionof{table}{Flow-head depth with frozen Kronos. After-fee return is in percent.}
\label{tab:depth}
\footnotesize\setlength{\tabcolsep}{2pt}
\begin{tabular}{@{}lrrr@{}}
\toprule
Depth / stage & Return & Sharpe & Calmar\\
\midrule
12L / IL &29.37&0.999&1.033\\
12L / RL &46.74&2.109&3.728\\
32L / IL &27.79&0.954&0.955\\
32L / RL &51.38&2.192&4.043\\
\bottomrule
\end{tabular}
\end{minipage}
\end{table}

\subsection{The Effect of Flow-head Size}
With the same frozen Kronos representation and two-stage training (Table~\ref{tab:depth}), a 12-layer head is slightly stronger than the 32-layer head after IL, both improve markedly under RL ($3.728$ Calmar for 12 layers), and the deeper head achieves the higher final return and Calmar.

\subsection{Reward choice and RL objective}
\label{sec:objective}
The group-relative update can only prefer the behavior its reward distinguishes, so Table~\ref{tab:reward} compares Sharpe, cumulative return, mean return, and a risk-adjusted-return reward against the IL starting policy on the same ledger. Cumulative return gives the highest annualized return, but the Sharpe-trained policy has the smallest MDD and highest Calmar while trading more than the return-focused variants, so its downside improvement is not simply the result of trading less; the risk-adjusted-return variant loses money, so nominal risk adjustment does not by itself ensure a useful training signal.

\rev{\paragraph{Deployment-aligned objectives.} Putting gross exposure or fees into the reward lowers the deployed performance (Appendix~\ref{app:closedloop}): scaling the Sharpe by gross exposure raises the MDD to $18.1\%$ and lowers the Calmar from $4.043$ to $2.76$, and charging fees lowers turnover to $1.30$ but raises the MDD to $26.0\%$ (Calmar $1.98$). With holdings as an additional input, the closed-loop policy reaches a Calmar of $4.59$ with a turnover of $1.38$, and a no-trade band at execution raises the Sharpe to $2.46$ and the Calmar to $5.75$.}

\begin{table}[ht]
\centering
\caption{\rev{Reward ablation.} Return and MDD are percentages.}
\label{tab:reward}
\footnotesize
\begin{tabular}{lrrrrr}
\toprule
Training reward & Return & MDD & Sharpe & Calmar & TO\\
\midrule
IL initialization &27.79&29.10&0.954&0.955&1.143\\
Sharpe &51.38&12.81&2.192&4.043&1.976\\
Cumulative return &60.62&16.79&2.150&3.611&1.658\\
Mean return &58.36&17.19&2.051&3.395&1.613\\
Risk-adjusted return &$-10.63$&28.60&$-0.480$&$-0.372$&2.070\\
\bottomrule
\end{tabular}
\end{table}

\rev{\paragraph{Policy gradients versus direct optimization.}\label{sec:draft} Back-propagating the same differentiable reward through the deterministic 4-step ODE from the same IL checkpoint (DRaFT; \citealt{clark2024draft}) reaches a Calmar of at most $1.74$ across four settings, versus $4.043$ for Flow-GRPO (Appendix~\ref{app:draft}). Policy gradients need only reward values, so they also apply to grid-rounded rewards such as the deployed P\&L; group-relative scoring standardizes and clips rewards within a group, whereas each pathwise gradient follows one realized price path and mainly raises turnover; and the stochastic sampler also provides the closed-form likelihoods that Flow-GRPO optimizes.}

\section{Conclusion}
\label{sec:conclusion}
\model\ turns raw multi-asset histories into complete signed portfolios through a cross-asset flow model: strategy-supervised imitation establishes structured, nontrivial trading actions, and group-relative RL with a frozen reference then improves their portfolio performance. Improvements across four demonstration priors support this progression, and architecture comparisons show that market representations must be paired with an action model that both expresses coordinated allocations and supports reward-based updates. Together, they realize a profitable instance of direct joint portfolio decision-making\rev{ that surpasses its teacher on every metric, holds on six further months of unseen data, and extends directly to closed-loop control}.

\clearpage

\bibliography{references}
\bibliographystyle{plainnat}

\clearpage
\appendix
\section{Mathematical Clarifications}
\subsection{A no-trade condition under uninformative expected returns}
\label{app:coldstart}
Consider a one-period decision from a zero book with conditional expected return vector $\mu\in\R^N$, proportional trading cost $\kappa\|w\|_1$, and risk penalty $\Psi(w)\geq0$ satisfying $\Psi(0)=0$. Let the feasible set contain zero, and define
\begin{equation}
 J(w)=\mu^\top w-\kappa\|w\|_1-\Psi(w).
\end{equation}
If $\|\mu\|_\infty\leq\kappa$, zero is a global maximizer of this objective. Indeed, H\"older's inequality gives
\begin{equation}
 J(w)\leq(\|\mu\|_\infty-\kappa)\|w\|_1-\Psi(w)\leq0=J(0).
\end{equation}
This elementary observation also holds after restricting the feasible set to dollar-neutral books. It motivates biasing initial portfolio proposals toward useful trading behavior.

A related optimization issue is that a group of identical zero books receives identical rewards, yielding zero group-relative advantages. The reward-driven surrogate then provides no directional signal for that group. This does not make zero an absorbing state in general: stochastic exploration, nonzero categorical probability mass, and reference regularization can still change a policy. The reported zero-book outcomes concern alternative action heads refined after IL\rev{; Section~\ref{app:init} reports the control without imitation}.

\subsection{Dollar neutrality, factor neutrality, and execution rounding}
\label{app:neutrality}
For $P=I-\one\one^\top/N$, $P^2=P$ and $\one^\top P=0$. Consequently, the continuous decoder in Equation~\ref{eq:decode} maps every latent to the dollar-neutral subspace. If asset returns follow $r=\beta f+u$, however, portfolio factor exposure is $w^\top\beta$. This need not vanish when $w^\top\one=0$. For example, $w=(a,-a)$ and $\beta=(2,1)$ give $w^\top\beta=a$. Dollar neutrality therefore cannot certify beta neutrality, sector neutrality, or bounded drawdown.

Rounding each position to a grid of width $q$ in allocation units can also introduce a residual net position. With fixed notional $A$, nearest-grid rounding gives the loose bound $|\one^\top\widehat w|\leq Nq/(2A)$, assuming no additional clipping. The recorded $A=10$ and $q=0.1$ correspond to $0.01$ increments in normalized portfolio weights. Single-name clipping in the teacher or factor pipeline may additionally alter neutrality. Exact deployed neutrality would require a final feasible projection, which is not assumed in the reported implementation.

\section{Implementation and Training Details}
\subsection{Market context and demonstration construction}
\label{app:context}
The reference universe chooses the top 20 assets by trailing 60-day traded amount and reconstitutes weekly on Sunday. The reference strategy first computes a signal for each asset, then applies its allocation pipeline: temporal smoothing over 24 hourly bars, volatility scaling with target $0.05$, a gross allocation cap of 10 units, a per-name clip of $\pm2$ units, and a $0.1$-unit execution grid. Demonstrations are the resulting allocations divided by fixed notional $A=10$.

\rev{\paragraph{AgenticAlpha.} AgenticAlpha is built only from training data. AlphaGen \citep{yu2023alphagen} and AlphaSAGE \citep{chen2026alphasage} each mine a set of alphas on the training period and combine them linearly. AgenticAlpha then combines the two methods at the portfolio level: the AlphaGen alphas and the AlphaSAGE alphas each manage half of the capital, and the two halves form one portfolio that passes through the allocation pipeline above. Table~\ref{tab:alphas} lists the five alphas with the largest absolute weights from each method; window lengths are in hourly bars, and the AlphaGen grammar writes them with the suffix \texttt{d}. We will release all the original alphas. Table~\ref{tab:teacher} compares the teacher with \model: imitation alone stays below the teacher, because it is trained to match the teacher's books rather than its return and serves as the starting distribution for RL, and RL then surpasses the teacher on every metric.}

\begin{table}[h]
\centering
\caption{The five AlphaGen and five AlphaSAGE alphas with the largest absolute weights in AgenticAlpha.}
\label{tab:alphas}
\resizebox{\linewidth}{!}{%
\begin{tabular}{llr}
\toprule
Source & Alpha expression & Weight\\
\midrule
AlphaGen & \texttt{Ref(Greater(\$vwap,-2.0),1d)} & $+0.1494$\\
AlphaGen & \texttt{WMA(Greater(Div(-5.0,Std(Greater(\$low,-0.5),48d)),-0.01),12d)} & $-0.1476$\\
AlphaGen & \texttt{Log(\$vwap)} & $+0.0207$\\
AlphaGen & \texttt{Less(2.0,Greater(Less(Greater(\$open,Mul(\$vwap,-2.0)),1.0),\$open))} & $-0.0049$\\
AlphaGen & \texttt{Sub(\$volume,Max(Greater(Div(5.0,\$volume),0.01),1d))} & $+0.0020$\\
\midrule
AlphaSAGE & \texttt{Div(Pow(Less(\$open,5.0),\$high),\$volume)} & $+0.0238$\\
AlphaSAGE & \texttt{TsMad(Add(-5.0,Mul(TsDiv(\$volume,12),10.0)),48)} & $-0.0229$\\
AlphaSAGE & \texttt{TsRank(Sub(-30.0,\$high),120)} & $+0.0218$\\
AlphaSAGE & \texttt{TsStd(TsDiv(Rank(Mul(0.01,Mul(Abs(\$vwap),-0.5))),120),120)} & $-0.0192$\\
AlphaSAGE & \texttt{Sign(Mul(-1.0,TsDelta(\$vwap,336)))} & $-0.0188$\\
\bottomrule
\end{tabular}}
\end{table}

\begin{table}[h]
\centering
\caption{Teacher, imitation, and refined policy on the same ledger with a 5 bps fee.}
\label{tab:teacher}
\small
\begin{tabular}{lrrrr}
\toprule
Policy & Return $\uparrow$ & MDD $\downarrow$ & Sharpe $\uparrow$ & Calmar $\uparrow$\\
\midrule
AgenticAlpha (teacher) & 33.92 & 19.57 & 1.472 & 1.733\\
\model\ after IL & 27.79 & 29.10 & 0.954 & 0.955\\
\model\ after IL + Flow-GRPO & \textbf{51.38} {\scriptsize$\pm$2.04} & \textbf{12.81} {\scriptsize$\pm$1.51} & \textbf{2.192} {\scriptsize$\pm$0.086} & \textbf{4.043} {\scriptsize$\pm$0.379}\\
\bottomrule
\end{tabular}
\end{table}

The model encoder uses window-normalized OHLCV and derived amount channels, with normalized values clipped to $[-5,5]$. The recent eight coarse and eight fine token states are pooled separately for each asset. Their summaries, a cross-asset market summary, asset identity embeddings, and calendar information condition the action head. The implementation reads encoder width from the loaded checkpoint configuration. We use the Kronos-base checkpoint in our implementation.

\subsection{Supervised auxiliary objectives}
\label{app:sft}
Let $h(u)$ denote the unit-threshold Huber loss: $h(u)=u^2/2$ for $|u|<1$ and $|u|-1/2$ otherwise. For a nonempty demonstrated book, define $\widehat w=P\widehat a/c$ and let $d_\theta(X)$ be the auxiliary direct-score output, with $w^{\mathrm{dir}}=Pd_\theta(X)/c$. The portfolio auxiliary terms are
\begin{align}
 \mathcal L_{\mathrm{weight}}&=\E\left[\frac1N\sum_j h(c\widehat w_j-cw_j^\star)\right],\\
 \mathcal L_{\mathrm{gross}}&=\E\left[h(\|c\widehat w\|_1-\|cw^\star\|_1)\right],\\
 \mathcal L_{\mathrm{score}}&=\E\left[\frac1N\sum_j h(cw_j^{\mathrm{dir}}-cw_j^\star)\right].
\end{align}
They supervise action-scaled weights, so their magnitude is comparable to the flow loss. Empty demonstrated books are excluded from these three auxiliary averages in the allocation-target path. The ranking term averages over pairs whose teacher weight gap exceeds $0.01$:
\begin{equation}
 \mathcal L_{\mathrm{rank}}=\E_{(i,j):|w_i^\star-w_j^\star|>0.01}
 \left[\log\!\left(1+\exp\{-\operatorname{sign}(w_i^\star-w_j^\star)
 (\widehat w_i-\widehat w_j)\}\right)\right].
\end{equation}
Flow time is sampled as $\tau=0.999(1-U)$ with $U\sim\operatorname{Beta}(1.5,1.0)$, and the velocity network uses 1,000 time buckets. These objectives encourage fidelity but do not force the neural policy to reproduce the teacher's temporal smoothness or executed turnover.

\subsection{Stochastic flow transitions and loss normalization}
\label{app:rl}
For $K$ transitions, let $\Delta=1/K$, $\tau_k=k/K$, and $s_k=1-\tau_k$, for $k=0,\ldots,K-1$. Set $d_k=\tau_k$ for $k>0$, and $d_0=1/K$. With noise coefficient $\nu=0.7$, the implemented stochastic flow sampler uses
\begin{align}
 \gamma_k&=\nu\sqrt{s_k/d_k},\\
 m_\theta(z_k,\tau_k)&=z_k\left(1-\frac{\gamma_k^2\Delta}{2s_k}\right)
 +v_\theta(z_k,\tau_k;X)\left(1+\frac{\gamma_k^2\tau_k}{2s_k}\right)\Delta,\\
 z_{k+1}&=m_\theta(z_k,\tau_k)+\gamma_k\sqrt\Delta\,\xi_k,
 \qquad \xi_k\sim\mathcal N(0,I).
\end{align}
All evaluated transitions occur before $\tau=1$, avoiding the terminal singularity. Setting $\nu=0$ recovers an Euler update. The training sampler uses ten stochastic transitions; deployed inference uses four deterministic Euler transitions conditional on a seeded initial Gaussian latent. Thus deterministic inference refers to repeatability for a given timestamp seed, not zero initial noise.

The scalar transition statistic is
\begin{equation}
 \ell_\theta=\frac1N\sum_{j=1}^N\left[
 -\frac{(z_{k+1,j}-m_{\theta,j})^2}{2\gamma_k^2\Delta}
 -\log(\gamma_k\sqrt\Delta)-\frac12\log(2\pi)\right].
\end{equation}
Consequently, $q=\exp(\ell_\theta-\ell_{\mathrm{old}})$ is the $N$th root of the full joint transition likelihood ratio. This dimension normalization changes the clipped surrogate and must be retained in a reproduction. It should not be interpreted as an exact untempered joint-policy likelihood-ratio estimator.

The two clipped surrogates in Equation~\ref{eq:rl} share the definition
\begin{align}
 \mathcal C(q,a)&=-\min\{qa,\clip(q,1-\delta,1+\delta)a\},\nonumber\\
 B_{i,k}&=\sg\!\left[\clip(\ell_{\mathrm S}^{i,k}-\ell_{\mathrm{old}}^{i,k},-5,5)\right],\qquad \delta=10^{-3}.
 \label{eq:clipped}
\end{align}
Here $\sg$ stops gradients through the reference advantage. Both surrogates use the same sampled transition and dimension-normalized ratio, but their advantages express different preferences: realized portfolio reward for $A_i$, and relative reference-policy support for $B_{i,k}$.

The separate reference term in Equation~\ref{eq:rl} is
\begin{equation}
 \mathcal K_k=\frac{\|m_\theta-m_{\mathrm S}\|_2^2}{2N\gamma_k^2}
 =\frac{\Delta}{N}D_{\mathrm{KL}}\!\left(
 \mathcal N(m_\theta,\gamma_k^2\Delta I)\,\|\,
 \mathcal N(m_{\mathrm S},\gamma_k^2\Delta I)\right).
\end{equation}
It is a scaled Gaussian divergence, because the denominator uses the diffusion coefficient rather than the full transition variance. The OPD surrogate uses a detached difference of teacher and old-policy log densities on the same sampled transition. This log-density-difference advantage follows the on-policy distillation pattern used in OpenClaw-RL \citep{wang2026openclaw}; here the reference is a frozen IL trading policy, without hindsight textual hints. The reference variant does not use an EMA teacher or noise-only optimization. Dropout is disabled during both stochastic rollout and likelihood recomputation, while gradients update the action head.

Training begins with historical contexts and causal strategy allocations. The frozen backbone encodes each context, and Equation~\ref{eq:il} trains the action head before a copy is retained as $\theta_{\mathrm S}$. Refinement samples eight old-policy stochastic paths per context and stores their states and mean log densities. Terminal portfolios are decoded, rescaled to gross $0.85$, and scored on the forward return window. After computing group-relative advantages, the trainable and reference heads recompute the sampled transition densities. Equation~\ref{eq:rl} averages over contexts, candidates, and ten transitions to update the action head.

Deployment retains this head but uses the shorter deterministic integration path. A timestamp seeds the initial Gaussian latent; four Euler steps generate the action, which is centered and divided by $c=10$. The rounded target enters the next ledger bar. Distinguishing the scoring and deployment paths matters: the former compares fixed candidate books over a local horizon, whereas the latter produces a sequence whose position changes incur fees.

\rev{\subsection{Closed-loop variant and deployment-aligned objectives}
\label{app:closedloop}
We start from one of the Flow-GRPO policies and give its action head the current holdings as an additional input. For each asset, six holdings features (the position, its signed logarithm, whether the asset is held, and the gross and net exposure of the book) are embedded by a two-layer MLP and added to the action token of that asset. The last layer of this MLP is zero-initialized, so training starts exactly from the trained policy. Training runs for 2000 steps with a learning rate of $10^{-5}$ and a batch size of 16. Table~\ref{tab:objective} compares this closed-loop variant and the deployment-aligned rewards of Section~\ref{sec:objective}.}

\begin{table}[h]
\centering
\caption{\textbf{Deployment-aligned objectives.} The alternatives are single runs with the same seed as the closed-loop variant; \model\ is reported as in Table~\ref{tab:main}.}
\label{tab:objective}
\small
\begin{tabular}{lrrrr}
\toprule
Training objective & MDD $\downarrow$ & Sharpe $\uparrow$ & Calmar $\uparrow$ & TO\\
\midrule
\model\ (gross-normalized Sharpe) & 12.81 & 2.192 & 4.043 & 1.976\\
Sharpe scaled by gross exposure & 18.1 & 1.76 & 2.76 & 2.07\\
Sharpe after fees & 26.0 & 1.37 & 1.98 & 1.30\\
Closed loop with holdings input & 11.9 & \textbf{2.15} & \textbf{4.59} & 1.38\\
\bottomrule
\end{tabular}
\end{table}

\rev{\subsection{Gaussian one-pass Transformer}
\label{app:gaussian}
The one-pass Transformer of Table~\ref{tab:action} is a bidirectional stochastic Transformer policy with about 1.0B parameters, matched to the flow head and reading the same tokens. It outputs a mean book $\mu$ and samples actions from $\mathcal N(\mu,\sigma^2I)$. Its stage 2 uses the same GRPO code as the flow head, with eight candidates per state, the same 24-bar gross-normalized Sharpe reward, the same clipping, KL, and distillation terms, and 2000 steps; the $\sigma=0.2$ run is the one reported in Table~\ref{tab:action}(a). The settings other than $\sigma=0.4$ trade about three times the book per day.}

\subsection{Recorded configuration and reproducibility boundaries}
\begin{table}[h]
\centering
\caption{Training and inference configuration.}
\label{tab:config}
\small
\setlength{\tabcolsep}{4pt}
\begin{tabular}{lp{0.61\linewidth}}
\toprule
Component & Recorded setting\\
\midrule
Market cadence / universe & Hourly; dynamic top 20; weekly reconstitution\\
Backbone & Frozen Kronos-base with its base tokenizer\\
Main flow head & 32 layers; 32 attention heads; head dimension 48; about 1.03B parameters\\
IL dropout / time sampling & 0.2; $0.999(1-U)$, $U\sim\operatorname{Beta}(1.5,1.0)$; 1,000 time buckets\\
Action scale / execution grid & $c=10$; grid 0.1 allocation units for notional 10\\
IL auxiliary coefficients & Score 1.0; weight 0.5; ranking 0.05; gross 0.1\\
RL sampler & 8 candidates; 10 stochastic transitions; noise coefficient 0.7\\
RL objective & Ratio clip $10^{-3}$; advantage clip 5; reference coefficient 0.004; OPD 0.1\\
Inference & 4 Euler steps; timestamp-seeded initial noise\\
Execution cost & 5 bps per absolute traded weight; net daily metrics\\
\bottomrule
\end{tabular}
\end{table}

\section{Accounting and Comparison Details}
\subsection{Ledger metrics and timing}
\label{app:metrics}
Let $\widehat w_t$ denote the grid-rounded target generated from bars available through $t$, with zero weight outside the eligible universe. In the reported lagged target-weight ledger, the target becomes the position for the following bar. A normalized expression for the accounting is
\begin{equation}
 g_{t+1}=\widehat w_t^\top r_{t+1},\qquad
 u_{t+1}=\|\widehat w_t-\widehat w_{t-1}\|_1,\qquad
 n_{t+1}=g_{t+1}-0.0005u_{t+1}.
\end{equation}
This describes the fixed-notional target-change convention\rev{: turnover is measured on changes of the target book}. For each day $d$, aggregate $n_d=\sum_{t\in d}n_t$, $u_d=\sum_{t\in d}u_t$, and let $D$ be the number of evaluated days. Then
\begin{align}
 \mathrm{AnnRet}&=\frac{365}{D}\sum_{d=1}^{D}n_d,&
 \mathrm{Sharpe}&=\sqrt{365}\,\frac{\overline n}{s(n)},&
 \mathrm{TO}&=\frac1D\sum_{d=1}^{D}u_d,\\
 C_d&=\sum_{j=1}^{d}n_j,&
 \mathrm{MDD}&=\max_{0\leq a\leq b\leq D}(C_a-C_b),&
 \mathrm{Calmar}&=\frac{\mathrm{AnnRet}}{\mathrm{MDD}},
\end{align}
with $C_0=0$. Metrics use zero as the risk-free benchmark. A zero-book run has no economically meaningful Sharpe or Calmar; our tables report both as No Trade (NT). All such entries are explicitly described as no-trade outcomes.

Native data alignment yields 423--425 evaluation days across methods. A common arithmetic metric convention does not equalize fills, rebalancing, or asset availability. Additive drawdown can exceed $100\%$ in this bookkeeping, as seen for some negative baselines; those values should not be read as feasible continuation of a funded account after insolvency.

Training rewards use fixed candidate weights over $H$ forward hourly returns. The code's after-cost variants compare the candidate with the model's deterministic previous-bar portfolio, computed without gradients, and penalize this change. This is different from replaying an entire future sequence of hourly decisions and their rebalancing costs.

\rev{\subsection{The 2026 out-of-sample period}
\label{app:oos}
Table~\ref{tab:oos} compares the original test with the new period from January to June 2026, evaluated with the same trained policy without retraining. The 2026 evaluation also settles funding for every contract on its own 4-hour or 8-hour schedule, using the funding history published by the exchange; on the original test, settling funding in this way keeps every metric within its variation across seeds.}

\begin{table}[h]
\centering
\caption{\textbf{Out-of-sample periods.} The 2026 period is evaluated with the same trained policy, without retraining.}
\label{tab:oos}
\footnotesize
\setlength{\tabcolsep}{3pt}
\begin{tabular}{lcccc}
\toprule
Out-of-sample period & Return $\uparrow$ & MDD $\downarrow$ & Sharpe $\uparrow$ & Calmar $\uparrow$\\
\midrule
Nov 2024--Dec 2025 (Table~\ref{tab:main}) & 51.38 {\scriptsize$\pm$2.04} & 12.81 {\scriptsize$\pm$1.51} & 2.192 {\scriptsize$\pm$0.086} & 4.043 {\scriptsize$\pm$0.379}\\
Jan--Jun 2026 (new) & \textbf{62.53} & \textbf{12.77} & \textbf{2.06} & \textbf{4.90}\\
\bottomrule
\end{tabular}
\end{table}

\rev{\subsection{Trading fees and bid-ask spread}
\label{app:costs}
The 5 bps fee is the maximum over the actual base-tier fees of major perpetual-futures venues, centralized and decentralized, and the fees used in prior crypto trading papers (Table~\ref{tab:fees}). Higher volume tiers, fee discounts (10\% when paying in BNB on Binance, up to 40\% for staking on Hyperliquid), and maker orders only lower the venue rates. Our backtest therefore uses the most conservative setting: it charges the base-tier taker fee on every change of the target book at every hourly rebalance, although a policy that trades toward a target book can execute with limit orders at a maker fee of at most 2 bps. All methods in the paper are evaluated with the same fee.

\model\ trades high-volume USD-margined perpetual contracts on Binance, re-selected every week by their trading volume over the past 60 days. These contracts have deep order books and are typically quoted at a spread of one tick. Half a tick corresponds to 0.24 bps of price for the median traded asset-hour, less than 5\% of the 5 bps fee charged on every unit of traded weight (0.44 bps when weighted by traded volume and 0.006 bps for BTC).}

\begin{table}[h]
\centering
\caption{Trading fees. Left: base-tier perpetual-futures fees of major venues. Right: fees per trade in published crypto trading papers.}
\label{tab:fees}
\footnotesize
\setlength{\tabcolsep}{3pt}
\begin{minipage}[t]{0.44\linewidth}
\vspace{0pt}\centering
\resizebox{\linewidth}{!}{%
\begin{tabular}{llrr}
\toprule
Venue & Type & Taker & Maker\\
\midrule
Binance USD-M & CEX & 5.0 & 2.0\\
OKX & CEX & 5.0 & 2.0\\
Hyperliquid & DEX & 4.5 & 1.5\\
Coinbase International & CEX & 4.0 & 2.0\\
\bottomrule
\end{tabular}}
\end{minipage}\hfill
\begin{minipage}[t]{0.54\linewidth}
\vspace{0pt}\centering
\resizebox{\linewidth}{!}{%
\begin{tabular}{llr}
\toprule
Paper & Market & Fee\\
\midrule
\citet{borrageiro2022rrl} & BitMEX perpetual, 5 min & 5\\
EarnHFT \citep{qin2024earnhft} & Binance, second-level & 2\\
MacroHFT \citep{zong2024macrohft} & Binance, minute-level & 2\\
ArchetypeTrader \citep{zong2026archetypetrader} & crypto long-short, 10 min & 2\\
FineFT \citep{qin2026fineft} & Binance perpetual, 5 min & 2\\
\model\ (ours) & crypto perpetual, 1 h & 5\\
\bottomrule
\end{tabular}}
\end{minipage}
\end{table}

\rev{\subsection{Pure-RL baselines}
\label{app:purerl}
We train three recent methods, MARS \citep{chen2026mars}, ReCAP \citep{pan2026recap}, and DeepAries \citep{kim2025deeparies}, and eight classic RL algorithms from scratch on our data. Each method uses its published architecture and hyper-parameters, adapted to our rotating universe, with the same data, split, fee, and ledger as \model. Methods that are long-only by design are kept long-only. For each classic algorithm we train long-only and long-short variants (19 agents in total) and report the better one, which is long-only in every case (Table~\ref{tab:purerl}).}

\begin{table}[h]
\centering
\caption{Pure-RL trading methods trained from scratch. Returns are annualized arithmetic returns on fixed notional, so very high turnover with the fee can exceed $-100\%$.}
\label{tab:purerl}
\small
\begin{tabular}{lrrr}
\toprule
Method & Return $\uparrow$ & MDD $\downarrow$ & Sharpe $\uparrow$\\
\midrule
MARS \citep{chen2026mars} & 25.9 & 52.3 & 0.50\\
DeepAries \citep{kim2025deeparies} & $-13.1$ & 89.1 & $-0.15$\\
ReCAP \citep{pan2026recap} & $-253.1$ & 409.5 & $-2.03$\\
DDPG \citep{lillicrap2016ddpg} & $-19.4$ & 83.9 & $-0.23$\\
PPO \citep{schulman2017ppo} & $-35.9$ & 106.3 & $-0.41$\\
A2C \citep{mnih2016a3c} & $-55.8$ & 129.2 & $-0.63$\\
SAC \citep{haarnoja2018sac} & $-59.3$ & 130.0 & $-0.67$\\
TD3 \citep{fujimoto2018td3} & $-80.5$ & 151.1 & $-0.91$\\
MADDPG \citep{lowe2017maddpg} & $-6.8$ & 45.3 & $-0.10$\\
MAPPO \citep{yu2022mappo} & $-48.3$ & 114.3 & $-0.58$\\
DQN \citep{mnih2015dqn} & $-76.7$ & 129.4 & $-1.06$\\
BTC buy-and-hold & 23.8 & 37.7 & 0.55\\
\midrule
\model\ (IL + Flow-GRPO) & \textbf{51.38} {\scriptsize$\pm$2.04} & \textbf{12.81} {\scriptsize$\pm$1.51} & \textbf{2.192} {\scriptsize$\pm$0.086}\\
\bottomrule
\end{tabular}
\end{table}

\rev{\subsection{Direct differentiable optimization}
\label{app:draft}
Table~\ref{tab:draft} lists the four DRaFT settings of Section~\ref{sec:draft}. The loss combines the Sharpe reward with an L1 anchor to the IL book and a penalty on gross exposure; all runs start from the same IL checkpoint, back-propagate the gross-normalized 24-bar Sharpe through the deterministic 4-step ODE for 2000 steps; Flow-GRPO is reported as in Table~\ref{tab:main}. Policy gradients are preferable for three reasons. First, Flow-GRPO only needs reward values, so it also applies to rewards computed on books rounded to the execution grid, such as the deployed after-fee P\&L, which provide no useful pathwise gradient. Second, each pathwise gradient follows a single realized 24-hour price path and pushes the book toward whatever that path rewarded, so a larger reward weight mainly raises turnover, to $3.59$ and $4.50$ per day in DRaFT 2 and 4; group-relative scoring instead standardizes rewards within each group and clips the advantage, which makes it robust to heavy-tailed returns. Third, the stochastic sampler that generates diverse candidate books also provides the closed-form likelihoods that Flow-GRPO optimizes.}

\begin{table}[h]
\centering
\caption{Direct differentiable optimization (DRaFT) versus Flow-GRPO from the same IL checkpoint.}
\label{tab:draft}
\footnotesize
\setlength{\tabcolsep}{4pt}
\begin{tabular}{lrrrrrr}
\toprule
Setting & Learning rate & Reward weight & Anchor weight & Calmar $\uparrow$ & Sharpe $\uparrow$ & TO\\
\midrule
IL start & -- & -- & -- & 0.955 & 0.954 & 1.143\\
DRaFT 1 & $10^{-5}$ & 0.1 & 1.0 & 1.14 & 1.07 & 1.17\\
DRaFT 2 & $10^{-5}$ & 1.0 & 1.0 & 1.04 & 0.87 & 3.59\\
DRaFT 3 & $5\times10^{-5}$ & 0.1 & 1.0 & 1.74 & 1.41 & 1.19\\
DRaFT 4 & $10^{-5}$ & 1.0 & 0.1 & 0.95 & 0.80 & 4.50\\
\midrule
Flow-GRPO (ours) & -- & -- & -- & \textbf{4.043} & \textbf{2.192} & 1.976\\
\bottomrule
\end{tabular}
\end{table}

\section{Additional Related Context}
\label{app:related}
The decision to generate a portfolio does not remove the established connections between forecasting, control, and execution. Formulaic alphas \citep{kakushadze2016alphas} remain useful interpretable signals and can supply strategy demonstrations. Dynamic trading theory explicitly models the cost of adjusting toward a target \citep{garleanu2013dynamic}, while deep hedging learns actions under trading constraints \citep{buehler2018hedging}. These connections highlight a distinction within direct action learning: choosing a target book and choosing the cost-aware path to that book are related but different problems. Our evaluated policy learns the former, and its sequence is assessed through a fee-charging ledger. \rev{Section~\ref{sec:objective} shows that holdings-conditioned control extends \model\ directly.}

\rev{Related preprints study adjacent problems.} Adversarial inverse-RL trading uses demonstrations from multiple behaviors \citep{orra2026learning}, and AlphaZeroBeta studies market-neutral portfolio RL \citep{belyakov2026alphazerobeta}. Generative portfolio research also uses diffusion to model market environments \citep{aghapour2025portfolio}, while decision-weighted flow matching targets distributions useful for downstream stochastic optimization \citep{xie2026dwfm}. These works motivate separating what is generated from how it is used. In \model, a sample is the joint signed action itself, and the same action model participates in both imitation and reinforcement. Our empirical analysis consequently examines which action model and strategy-guided search configuration realize profitable joint decisions.


\rev{\section{Terminology}
\label{app:terms}
Table~\ref{tab:terms} defines the financial terms used throughout the paper.

\begin{table}[h]
\centering
\caption{Terminology.}
\label{tab:terms}
\small
\begin{tabular}{lp{0.62\linewidth}}
\toprule
Term & Definition\\
\midrule
Bar & Market data aggregated over one fixed time interval; \model\ uses 1-hour bars.\\
OHLCV & The open, high, low, and close prices and the traded volume within a bar.\\
Kronos & A financial foundation model \citep{shi2025kronos} pretrained to predict the next bar from tokenized OHLCV sequences; \model\ keeps it frozen and uses its hidden states as features.\\
Perpetual future & A futures contract without expiry that can be held long or short; holders of long and short positions exchange periodic funding payments.\\
Long-short target book & The signed weights over the eligible assets that the policy outputs every hour; positive weights are long positions and negative weights are short positions.\\
Gross exposure & The sum of absolute weights, $\|w\|_1$.\\
Turnover (TO) & The average daily sum of absolute weight changes.\\
Basis point (bps) & $0.01\%$; a 5 bps fee costs $0.05\%$ of the traded value.\\
Sharpe ratio & Annualized mean of daily net returns divided by their standard deviation.\\
Maximum drawdown (MDD) & The largest peak-to-trough loss of cumulative net return.\\
Calmar ratio & Annualized net return divided by MDD.\\
Teacher strategy & The strategy whose portfolios the action head imitates in stage 1.\\
\bottomrule
\end{tabular}
\end{table}
}

\end{document}